\documentclass[11pt,draftcls,onecolumn,journal]{IEEEtran}
\usepackage{amsmath,amsfonts}
\usepackage{algorithmic}
\usepackage{algorithm}
\usepackage{array}
\usepackage[caption=false,font=normalsize,labelfont=sf,textfont=sf]{subfig}
\usepackage{textcomp}
\usepackage{stfloats}
\usepackage{url}
\usepackage{verbatim}
\usepackage{graphicx}
\usepackage{cite}
\usepackage{fontawesome}
\usepackage{hyperref}
\usepackage{booktabs}
\usepackage{booktabs}   
\usepackage{tabularx}    
\usepackage{graphicx}    
\usepackage{multirow}    
\usepackage{xcolor}      
\usepackage{tikz}
\usepackage{threeparttable}
\begin{document}

\title{Brain Foundation Models: A Survey on
Advancements in Neural Signal Processing and
Brain Discovery}

\author{Xinliang~Zhou$^{\dag}$,~\IEEEmembership{}
        Chenyu Liu$^{\dag}$,~\IEEEmembership{}
        Zhisheng Chen,~\IEEEmembership{}
        Kun Wang,~\IEEEmembership{}
        Yi Ding$^{\ast}$,~\IEEEmembership{}
        Ziyu Jia$^{\ast}$,~\IEEEmembership{}
        and Qingsong Wen$^{\ast}$~\IEEEmembership{Senior~Member~IEEE}
\thanks{$^{\dag}$Equally Contribution $^{\ast}$Corresponding Authors}
\thanks{Manuscript received March 1, 2025, revised June 1, 2025, and accepted on July 15, 2025. }
\thanks{}
}

\markboth{}%
{}

\maketitle

\vspace{-95pt}
\section*{}
\label{sec:abstract}
{\color{black}

Brain foundation models (BFMs) represent a transformative paradigm in computational neuroscience that leverages large-scale pre-training on diverse neural signals to achieve robust generalization across tasks, modalities, and experimental contexts. This survey establishes the first comprehensive definition and framework for BFMs, systematically examining their construction, core methodologies, and applications. We present key approaches for data processing and training strategies alongside diverse applications spanning brain decoding and scientific discovery. Through critical analysis of recent methodological innovations, we identify fundamental challenges that must be addressed to realize the full potential of BFMs, including advancing data quality and standardization, optimizing model architectures, improving training efficiency, and enhancing interpretability. By bridging the gap between neuroscience and artificial intelligence, BFMs present unprecedented opportunities to revolutionize brain research, clinical diagnostics, and therapeutic interventions. This survey serves as a foundational reference for researchers and practitioners seeking to understand and advance this emerging field.
}

\section{Introduction}
\label{sec:scope}

\begin{figure}
    \centering  \includegraphics[width=0.8\linewidth]{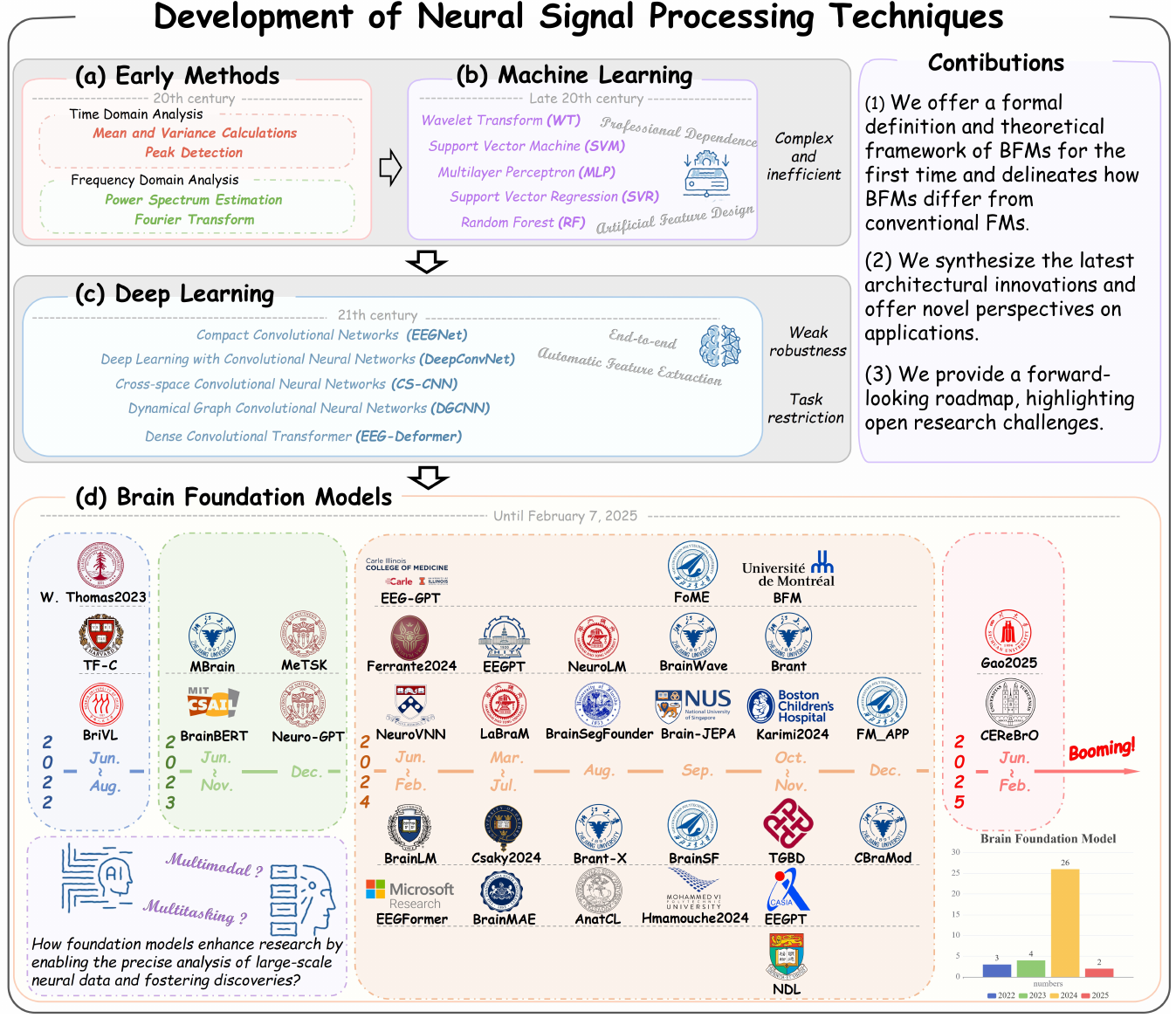}
    \caption{\textcolor{black}{\scriptsize Evolution of neural signal processing from traditional methods to BFMs. Panels (a) to (c) illustrate the progression from statistical techniques (e.g., PCA, ICA) and machine learning (e.g., SVM, KNN) to deep learning (e.g., CNN, LSTM) and BFMs. BFMs enhance the precision and scope of large-scale neural data analysis, enabling breakthroughs in neuroscience. Since 2022, BFMs like BrainLM and LaBraM have driven rapid advancements and methodological diversity in brain research.}}
    \label{fig:Development}
\end{figure}

Neural signal processing has advanced dramatically over the past century, driven by progress in neuroscience and computational methods, as shown in Fig. \ref{fig:Development}. Early research focused on brain recordings to study sleep, epilepsy, and cognition, leading to the adoption of statistical techniques like principal component analysis (PCA) \cite{cichocki2015tensor} and independent component analysis (ICA) \cite{kachenoura2008ica} for reducing dimensionality and isolating meaningful signal components amidst noisy environments. By identifying core features hidden within linearly-structured neural data,  statistical methods facilitated significant breakthroughs in both basic science and clinical applications, for example, identifying distinct electroencephalogram (EEG) frequency bands and distinguishing epileptic spike patterns. Traditional machine learning models, such as support vector machines (SVMs), random forests, and k-nearest neighbors (KNN) \cite{wu2023signal}, further advanced these developments by providing more robust classification and regression frameworks. They addressed challenges ranging from automated seizure detection to more applications, such as distinguishing emotional states from EEG. However, traditional machine learning methods were limited by their reliance on handcrafted features, sensitivity to variations in data quality across recording setups, and limited ability to model the nonlinear and dynamic nature of brain activity. As experimental paradigms became more diverse, spanning motor imagery, visual evoked potentials, and cognitive load tasks, the shortcomings of manual feature engineering and the poor adaptability of conventional machine learning methods became increasingly evident.

The rise of deep learning in the early 2010s revolutionized neural signal processing by enabling end-to-end feature learning directly from raw data. Deep neural networks, which can automatically learn useful features from raw or minimally processed signals, provide a powerful alternative to labor-intensive feature extraction pipelines. Recurrent models, especially long short-term memory (LSTMs), improved temporal modeling of neural signals \cite{ramachandram2017deep}. This capability proved highly beneficial for brain-computer interfaces (BCI), where time-dependent patterns, such as oscillatory rhythms or phase-locked responses, are essential for decoding user intentions or mental states. Convolutional neural networks (CNNs), with their ability to learn spatial and temporal filters, demonstrated unprecedented capabilities in simultaneously decoding both spatial and temporal patterns \cite{holobar2021noninvasive}. This led to state-of-the-art performance across tasks as diverse as motor imagery classification, steady-state visual evoked potential (SSVEP) detection, and cognitive workload assessment. Recently, Transformer architectures have been explored in BCI, given their aptitude for capturing long-range dependencies in sequential data \cite{chen2025emerging}. By processing input signals in parallel, rather than strictly sequentially, Transformers offer enhanced scalability and the potential for integrated multimodal analysis. However, early deep learning models were still constrained by their relatively narrow specialization. Researchers often had to adapt or craft network architectures for specific tasks, such as speech-evoked EEG decoding and motor imagery, limiting their applicability across diverse paradigms and data types. Furthermore, this task-centric design hindered generalization to data collected under heterogeneous conditions, including varied sensor setups, subject populations, and stimulus protocols.

These challenges sparked increasing interest in foundation models (FMs)\cite{bommasani2021opportunities}, large-scale systems pretrained on broad data distributions to learn universal representations. The remarkable success of FMs in natural language processing (NLP) and computer vision (CV), exemplified by models like BERT \cite{devlin2019bert}, GPT \cite{radford2018improving}, and Segment Anything (SAM) \cite{kirillov2023segment}, demonstrated the power of leveraging large, diverse datasets and self-supervised objectives to capture complex statistical patterns. By pretraining on text, images, or video from various sources, these models learned generalized features that could be adapted to a wide array of downstream tasks, often outperforming traditional models in similar applications. Inspired by this, researchers began exploring the potential of FMs for neural signals, hypothesizing that their ability to capture complex dependencies could benefit EEG and functional magnetic resonance imaging (fMRI) analysis. With pretraining techniques like masked signal modeling, FMs showed promise for significant performance improvements in tasks like mental state classification, motor imagery, and neurodegenerative disease diagnosis. However, directly applying these FMs to neural data presents fundamental challenges. Compared to text or images, neural signals exhibit greater spatiotemporal complexity and often have a lower signal-to-noise ratio (SNR). Additionally, recordings can vary significantly across individuals and are often subject to strict constraints, including patient privacy and institutional review protocols. Moreover, effective brain research and clinical translation require simultaneous support for decoding, diagnosis, and dynamic network modeling, an integrated approach that extends beyond the focus of most existing FMs.

To bridge this gap, Brain Foundation Models (BFMs) offer a paradigm that unifies neural signal processing through large-scale, neurocentric pretraining and cross-scenario, cross-task capabilities. Unlike general-purpose FMs, which typically rely on massive text or image corpora, BFMs aim to incorporate large-scale datasets of neural signals, such as EEG or fMRI recordings spanning thousands of subjects and numerous hours. The BFMs approach is underpinned by three guiding principles: (1) pretraining tailored to the dynamics of neural data, (2) enabling zero- or few-shot generalization across distinct scenarios, tasks, and modalities, reflecting the broad variability of brain states and experimental conditions in neuroscience, and (3) embedding AI mechanisms that address the sensitive nature of human neural recordings, such as federated learning frameworks to protect data privacy and rigorous anonymization protocols to safeguard participant identities.

This survey presents the first survey of BFMs and highlights the following contributions:

\begin{itemize}
    \item First, we offer a \textbf{formal definition and theoretical framework of BFMs for the first time} and delineates how BFMs differ from conventional FMs, underscoring the specialized data structures and objectives pertinent to neural signals.

    \item Second, we synthesize the \textbf{latest architectural innovations and offer novel perspectives on their applications.} Our analysis demonstrates how BFMs can extend the frontiers of brain decoding by achieving robust generalization across scenarios and tasks as well as brain simulation and discovery by enabling the construction of digital twin brains, respectively.

    \item  Finally, we provide a \textbf{forward-looking roadmap}, highlighting open research challenges, including data integration and quality, training strategies, multimodal architectural design, interpretability and regulatory considerations that must be addressed to fully realize the potential of BFMs.
\end{itemize}

 Taken together, these efforts promise to reshape the landscape of neural signal processing, charting a path toward more robust, versatile, and ethically responsible models that can adapt to the complexities and nuances of the human brain.






\section{Brain Foundation Models}

\subsection{Definition and Framework of BFMs}

\subsubsection{Definition}
BFMs refer to foundation models built using deep learning and neural network technologies pretraining on large-scale neural data designed to decode or simulate brain activity. These models aim to capture and understand the complex patterns in neural signals, thereby advancing neuroscience exploration and enabling brain disease diagnosis and treatment. BFMs integrate multimodal brain signal processing (e.g., EEG and fMRI), biological principles, and artificial intelligence techniques to extract deep neural activity patterns from large-scale data and multidimensional features. This allows for a multi-angle and precise interpretation of brain function.

{\color{black}
BFMs can be classified into three categories based on their training and application strategies. \textbf{Pretrained-only models} are constructed by pretraining on large-scale brain signal datasets to develop a robust feature extraction capability, which can be applied to general brain activity analysis and disease screening. After pretraining, these models are ready for deployment without the need for further fine-tuning. \textbf{Pretrained with fine-tuning models} are pretrained on extensive brain signal data and then fine-tuned for specific applications. This approach is used for tasks such as brain disease diagnosis, cognitive state assessment, and other specialized applications. \textbf{Pretrained with interpretability for brain discovery models} combine pretraining with interpretability techniques, such as perturbation analysis, to simulate and explore key biological mechanisms in brain activity. This goal is not only to provide high-precision diagnostic support but also to drive brain discovery by analyzing the neural underpinnings of brain functions.
}

\begin{figure}
    \centering
    \includegraphics[width=0.75\linewidth]{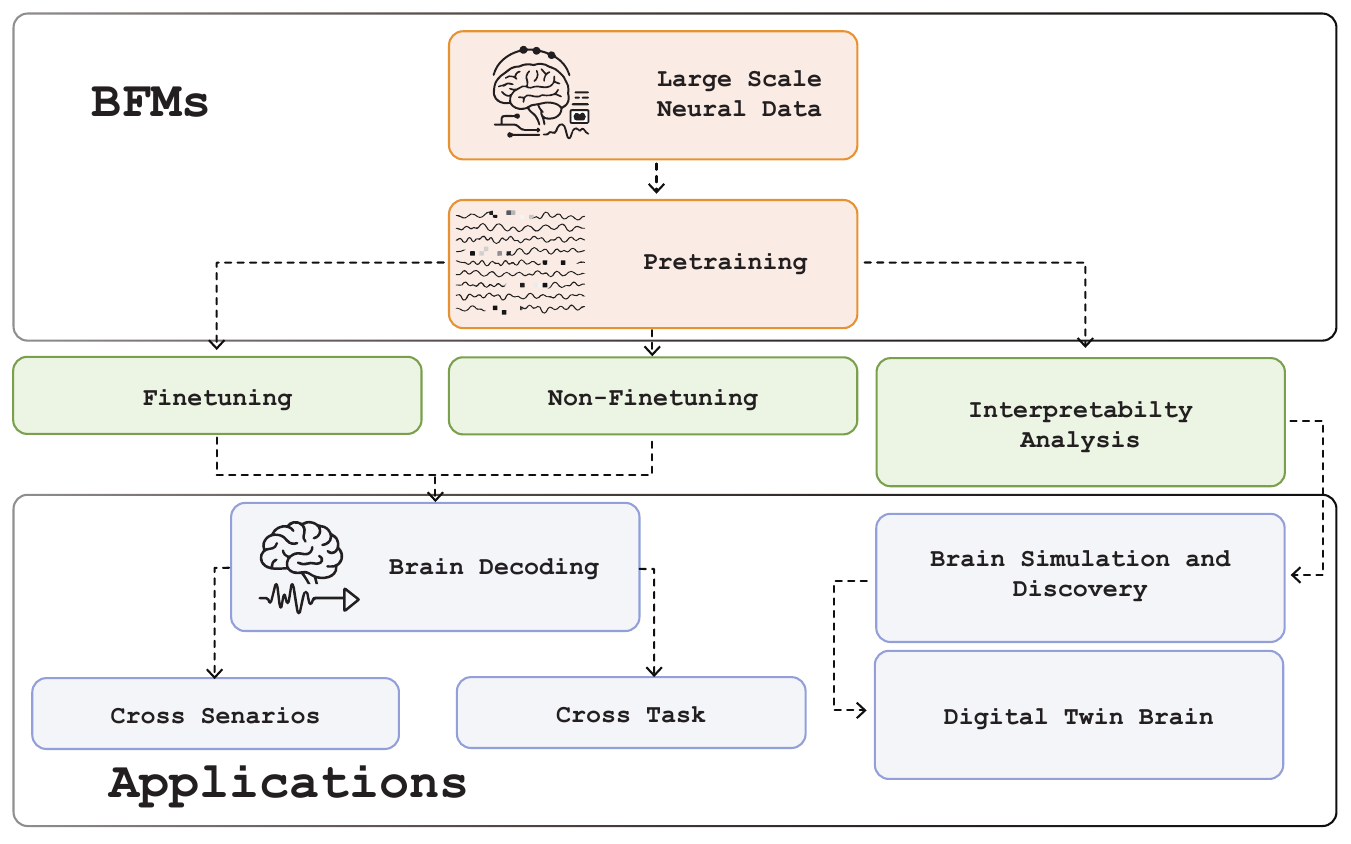}
    \caption{\textcolor{black}{\scriptsize This framework illustrates the key components involved in developing BFMs and their diverse applications. It emphasizes the use of large-scale neural data, pretraining (with or without fine-tuning), and interpretability analysis. The framework supports both brain decoding and simulation, with applications that span across cross-scenario and cross-task domains, as well as the creation of digital twin brain for brain discovery.}}
    \label{fig:framework}
\end{figure}

\subsubsection{Framework}
The BFMs framework, as shown in Fig. \ref{fig:framework}, provides a clear view of how BFMs and their applications are constructed. The framework is structured around three key components: large-scale neural data, pretraining (with or without fine-tuning), and interpretability analysis. These components work synergistically to enable BFMs to decode and simulate brain activity across different scenarios and tasks.

\begin{itemize}
\item \textbf{Large-Scale Neural Data:} At the core of the BFMs framework is the utilization of large-scale datasets of neural signals, including EEG, fMRI, and other brain activity recordings. These datasets are crucial as they provide the necessary diversity and volume to allow the model to learn generalized representations of brain function. The data often span across thousands of subjects and hours of brain activity, helping the model identify universal patterns that are applicable to a broad range of brain states and conditions. A larger, more varied dataset improves the BFM’s ability to generalize across experimental conditions, which is crucial for real-world applications.
\item \textbf{Pretrained with or without Fine-tuning:} 

The BFMs framework allows for two paths in model training: pretraining with or without fine-tuning.

\begin{itemize}
    \item \textbf{Pretraining with Fine-tuning:} These models are pretrained on extensive brain signal data and then fine-tuned for specific applications. Fine-tuning is often essential for tasks like brain disease diagnosis, cognitive state assessment, or task-specific BCI applications (e.g., motor imagery or emotion detection). It enhances performance by adapting the model to the nuances of each task.
    \item \textbf{Pretraining without Fine-tuning:} For some BFMs, pretraining alone is sufficient, allowing direct application without fine-tuning. These models utilize in-context learning, adapting to new scenarios or data through exposure during inference. In such cases, the model leverages its generalizable knowledge to perform tasks effectively without additional task-specific training.
    
\end{itemize}

\item \textbf{Interpretability Analysis:}  Interpretability techniques, such as perturbation analysis, attention mechanisms, and saliency mapping, allow researchers to investigate which parts of the brain contribute most to the model's predictions. This is particularly valuable in brain discovery applications, where the goal is to understand the underlying neural mechanisms that drive cognitive processes or disease states. By simulating and perturbing the digital brain created during model training, researchers can gain insights into the biological brain's functioning and explore unknown aspects of brain dynamics and disease mechanisms.
\end{itemize}

\subsection{Key Differentiators from Conventional Foundation Models}
\begin{table}[ht]
\centering
\caption{\scriptsize BFMs vs. Traditional Foundation Models}
\vspace{-10pt}
\label{tab:comparison}
\resizebox{0.8\linewidth}{!}{
\begin{tabularx}{\linewidth}{@{}lXX@{}}
\toprule
\textbf{Aspect} & \textbf{BFMs} & \textbf{Traditional FMs (e.g., LLMs, Vision FMs)} \\
\midrule
Application Domain & (1) Computational neural sciences; & (1) Natural language processing; \\ & (2) Neural simulations; & (2) Computer vision; \\ &(3) Human-AI interaction. & (3) Multimodal tasks. \\

Data Characteristics & (1) High noise; & (1) Text corpora; \\ & (2) Non-stationary signals; & (2) Image datasets; \\ &(3) Heterogeneous data alignment. & (3) Multimodal datasets. \\

Learning Objectives & (1) Universal neural expression; & (1) Semantic understanding; \\ & (2) Spatial and temporal modeling; & (2) Image generation; \\ &(3) Neuroscience constraints. & (3) Token prediction. \\

Ethical Constraints & (1) Privacy of neural data; & (1) Mitigating biases;  \\ & (2) Biological interpretability; & (2) Reducing mirage. \\ & (3) Clinical safety. \\

Output & (1) Predictions and simulation of actions; & (1) Text generation; \\ & (2) Decisions; & (2) Image synthesis; \\ &(3) Interactions in neural activities. & (3) Multimodal outputs. \\

\bottomrule
\end{tabularx}}
\label{tab:differ}
 \end{table}

Although both BFMs and traditional FMs serve as foundational models in their respective domains, they differ significantly in many ways. The key differentiators are summarized in the Tab. \ref{tab:differ}. BFMs are specifically designed to handle the complexities of neural data, enabling precise analysis and fostering discoveries in neuroscience. These models enable robust generalization across diverse scenarios, tasks, or modalities, thus playing a crucial role in advancing brain research. Besides, the pretraining for neural data highlights the unique role of BFMs in advancing brain research. However, traditional FMs, such as LLMs and Vision FMs, are adept at processing text and images, which propel advancements in language and vision applications. They are primarily applied in natural language processing, computer vision, and multimodal tasks to deal with text corpora, image datasets, and multimodal datasets.

Specifically, existing foundation models, such as BERT, GPT, and SAM, have demonstrated remarkable abilities within their respective domains by leveraging extensive pretraining on large-scale textual or visual corpora. BERT employs bidirectional contextualization to learn deep language representations, while GPT uses autoregressive language modeling to generate coherent text and exhibits strong zero-shot learning abilities. SAM, on the other hand, is good at universal image segmentation through prompt-based interaction. In contrast, BFMs are specifically designed to process high-noise, non-stationary neural signal data, such as EEG and fMRI recordings. They focus on extracting universal neural representations to achieve spatial and temporal modeling within the limits of neuroscience constraints. In addition, following standardized ethical constraints regarding neural data privacy and clinical safety is another direction of research for BFMs. This distinction highlights the unique challenges and opportunities BFMs present in advancing neuroscience research and applications.

\subsection{Methodological Components of BFMs}

Current BFMs can be broadly categorized into two key methodological components: Data Processing and Training Strategies. Data Processing focuses on addressing structural inconsistencies in neural recordings, such as heterogeneous channel configurations and variations in signal length across datasets, ensuring compatibility and stability for downstream modeling. However, strategies define how BFMs learn meaningful representations from brain signals, primarily through reconstruction-based and contrastive learning, each capturing different aspects of neural activity.

\subsubsection{Data Processing} It is a crucial part for BFMs to ensure consistency across diverse neural datasets. However, two major challenges arise.

One of the key challenges in data processing for BFMs is the structural heterogeneity of channel configurations and the inconsistency in signal length across datasets. Due to the lack of standardized electrode placements, brain signal data exhibit significant spatial inconsistencies. A channel, representing an electrode that captures neural activity, varies in number, spatial distribution, and recording characteristics across studies. For example, the DEAP dataset \cite{koelstra2011deap} uses a Biosemi 32-channel system, while the SEED dataset \cite{zheng2015investigating} employs a Neuroscan 62-channel system, leading to differences in both channel count and electrode placement. Since each channel encodes distinct neurophysiological information, simple interpolation or resampling may distort critical neural features, complicating cross-dataset integration.

Additionally, signal length inconsistencies arise from variations in experimental paradigms, as different cognitive and physiological processes unfold on distinct timescales. Neither excessively long nor short recording windows accurately capture neural activity. For instance, steady-state visual evoked potentials (SSVEPs) elicit responses within milliseconds, requiring brief recordings, whereas motor imagery tasks necessitate signal durations of several seconds. Sleep studies extend this further, spanning hours to capture full sleep cycles, while emotion-related studies not only involve prolonged signal acquisition but also exhibit diverse temporal characteristics, encompassing stable states, transient fluctuations, and sustained effects over varying timescales.

\textbf{Methodologies:}
Current BFMs adopt a discretized signal segmentation approach to address inconsistencies in data structure. The core principle of this method is to partition continuous brain signals into standardized time segments, ensuring a uniform input structure for the model. Given a raw brain signal represented $X \in \mathbb{R}^{C \times T}$, where $C$ denotes the number of channels and $T$ represents the total time steps, BFMs apply a fixed time window $w$ for segmentation while incorporating a sliding step $s$ to generate consecutive segments, formulated as:
\begin{equation}
x = \{x_{c, k} \in \mathbb{R}^{w} \mid c = 1, 2, ..., C; k = 1, 2, ..., \lfloor \frac{T - w}{s} \rfloor + 1 \}
\end{equation}
where each $x_{c, k}$ represents a fixed-length signal segment extracted from channel $c$ at time step $k$. Consequently, the total number of segmented patches is $|x| = C \times \left( \lfloor \frac{T - w}{s} \rfloor + 1 \right)$. This segmentation strategy enables BFMs to handle channel-wise variability by independently partitioning each channel while also unifying signal durations across tasks.

However, while discretized signal segmentation standardizes data formats, it inevitably leads to the loss of positional information. The segmented fragments lack their original spatiotemporal context within the neural signals, making it challenging for the model to capture cross-temporal and cross-regional dynamics. For instance, audiovisual integration relies on synchronized activity between the temporal and parietal lobes \cite{setti2023modality}. To compensate for this loss of spatiotemporal information, positional encoding is introduced to help the model correctly interpret the relative positions of signal segments in both the temporal and channel dimensions. Currently, BFMs employ two main strategies for utilizing positional encoding: fixed positional encoding and learnable positional encoding. The former uses a standard sinusoidal function to provide temporal information without trainable parameters. For example, Brain-JEPA \cite{dong2025brain} employs a positional embedding matrix based on brain gradient positioning for brain regions and a temporal encoding matrix using sine and cosine functions for temporal positioning.
The latter uses trainable parameters to learn optimal positional representations, enabling the model to adapt to different datasets and capture specific spatiotemporal patterns. For example, Brant \cite{zhang2023brant} defines a learnable positional encoding matrix $\mathbf{W}_{\text{pos}} \in \mathbb{R}^{L \times D}$ to capture the temporal order of patches, where $L$ represents the number of positional tokens, and $D$ denotes the feature dimension. This encoding is added to the projected input, allowing the model to learn sequence dependencies effectively. LaBraM \cite{jianglarge} extends this approach by introducing a learnable temporal encoding matrix $ \mathbf{W}_{\text{temp}} \in \mathbb{R}^{L \times D}$ and a spatial encoding matrix $\mathbf{W}_{\text{spat}} \in \mathbb{R}^{|C| \times D}$.

{\color{black}
\textbf{Remark:} \textit{Unlike foundational models in other domains, brain data exhibits differences in position encoding across various dimensions. In the temporal dimension, neural signals exhibit relatively stable temporal patterns, making fixed-position encoding more suitable to enhance generalization and avoid overfitting. In the spatial dimension, functional connectivity and topological structures between brain regions vary significantly across individuals and task contexts, making learnable position encoding more appropriate to accommodate diverse spatial relationships and task-related patterns. Building on this, the rotation-based position encoding adopted by current mainstream foundational models offers a solution that balances stability and expressive power. It retains the advantages of fixed encoding while introducing relative position modeling capabilities, making it suitable for modeling long-term dependencies in neural signals.}

\textbf{Summary of Data Processing:}
Data processing for BFMs is still in its early stages and faces challenges that limit its adaptability. The fixed time window strategy often does not match the varied timing of neural activity or individual differences, which can introduce biases and reduce model robustness. Managing inconsistent brain region configurations across datasets is another major issue; methods like regional intersections (e.g., CBRAMOD \cite{wang2024cbramod}) can cause information loss, while unions (e.g., EEGPT \cite{wang2025eegpt}) can increase the computational load. These inflexible standardization techniques, along with treating brain regions as separate channels without considering their structural relationships, also hinder biologically meaningful learning. Therefore, there is a strong need for more flexible and adaptive representations that can manage variations in temporal structure and brain region configurations while keeping essential physiological information.
}
\subsubsection{Training Strategy} It is a fundamental component of BFMs, as it determines how models extract meaningful representations from complex and noisy neural signals. Given the high variability, individual differences, and task-dependent dynamics inherent in brain data, effective training paradigms are essential to enable BFMs to generalize across diverse experimental conditions.

Most BFMs follow a two-stage training paradigm: pretraining and fine-tuning. During pretraining, models typically leverage self-supervised learning (SSL) to learn generalized neural signal representations from large-scale and diverse datasets without requiring labeled data. Some BFMs are then fine-tuned on specific downstream tasks using supervised learning, while others are designed for direct application post-pretraining (pretrained-only models, such as NeuroLM\cite{jiang2024neurolm} ).

\textbf{Methodologies:} 
SSL is central to the pretraining phase and operates by defining proxy tasks that enable models to learn useful features from unlabeled data. In BFMs, SSL objectives primarily fall into two categories: \textbf{reconstruction-based learning}, where models learn to recover missing parts of the input from the available context, and \textbf{contrastive learning}, where models learn to distinguish between similar and dissimilar pairs of brain signals.

{\color{black}
Formally, let $x_i \in \mathbb{R}^d$ represent the input brain signal, where $d$ denotes the dimensionality of the data (for example, the number of EEG channels). The goal is to learn a function $f$ such that the learned representation $f\left(x_i\right)$ captures the underlying patterns in the signal. A common SSL objective is to minimize the reconstruction error or to maximize the similarity between different parts of the input data, which is expressed mathematically as:

\begin{equation}
\mathcal{L}_{\text {SSL }}=\alpha \sum_i \mathcal{L}_{\text {reconstruction }}\left(x_i, \hat{x}_i\right)+\beta \mathcal{L}_{\text {contrastive }}\left(f\left(x_i\right), f\left(x_j\right)\right)
\end{equation}
where $\hat{x}_i$ represents the model's prediction for $x_i$, and $\mathcal{L}_{\text {reconstruction }}$ is a reconstruction loss (e.g., Mean Squared Error), which ensures that the model learns to predict missing parts of the signal. The contrastive loss $\mathcal{L}_{\text {contrastive }}$ is used to make the model learn similar representations for similar inputs and dissimilar representations for different ones. The coefficients $\alpha \in \{0, 1\}$ and $\beta \in \{0, 1\}$ act as indicators to select the desired loss components, where $(\alpha, \beta) \neq (0,0)$.

\begin{itemize}
    \item If $\alpha=1$ and $\beta=0$, the SSL objective focuses solely on reconstruction.
    \item If $\alpha=0$ and $\beta=1$, the SSL objective focuses solely on contrastive learning.
    \item If $\alpha=1$ and $\beta=1$, both reconstruction and contrastive learning objectives are jointly optimized.
\end{itemize}

Reconstruction-Based Learning: Its core principle is to mask or corrupt a portion of the input signal and require the model to recover the missing part, thereby learning the intrinsic structure of the data. This approach is widely applied in brain signal modeling as it effectively extracts both local and global features from temporal sequences, enhancing the model’s understanding of neural activity. Reconstruction-based methods can generally be divided into two main types:  \textbf{masked prediction} and \textbf{autoregression}. Masked prediction involves randomly masking segments of the neural signal, training the model to reconstruct the missing information based solely on the unmasked portions. Autoregression constrains the model to rely exclusively on historical data, enabling it to predict future neural signals in a sequential manner.


Masked prediction aims to learn global representations from incomplete input data. From a neuroscience perspective, this approach aligns with the redundant coding principle of the brain, wherein neural activity in certain regions can often be inferred from surrounding information. For instance, EEG signals exhibit strong temporal correlations, allowing masked time segments or channels to be reconstructed from the remaining data. If $X$ represents the input brain signal segments and $\tilde{X}$ is the version with certain segments masked (denoted by a set of masked indices $M$), and $f_{\theta}$ is the BFM with parameters $\theta$ that attempts to reconstruct the masked segments $\hat{X}_{masked}$ from the unmasked segments $X_{unmasked}$, the objective is often to minimize a reconstruction loss. For example, using Mean Squared Error (MSE) for continuous signals, this can be formulated as:
\begin{equation} \label{eq:recon_loss}
L_{\text{recon}} = \frac{1}{|M|} \sum_{i \in M} (X_i - \hat{X}_i)^2
\end{equation}
where $X_i$ is the original value of a masked element and $\hat{X}_i$ is its predicted value by the model $f_{\theta}(X_{\text{unmasked}})$. The specific nature of $X$ (e.g., raw signal values, spectral features) and the masking strategy can vary. 

In this context, masking can be performed in various ways, depending on the task at hand:

\begin{itemize}
    \item Temporal Masking: Specific time windows in the neural signal (e.g., EEG or fMRI) are masked. The model learns to predict the missing segments from surrounding temporal information. This is particularly useful in modeling dynamic brain activity over time.
    \item Spatial Masking: In fMRI or EEG data, certain spatial regions (such as brain regions or channels) are masked, and the model is tasked with reconstructing or predicting the activity in these regions based on the remaining parts of the signal.
\end{itemize}
}

For example, BrainBERT \cite{wang2023brainbert} adopts a time-frequency masking strategy, where randomly selected segments of the EEG spectrograms are obtained by the short-time Fourier transform or the Superlet transform and subsequently reconstructs them. It employs a Transformer-based architecture, leveraging self-attention mechanisms to extract contextual information from the unmasked regions and predict the missing spectral features. Brain-JEPA further refines this masking strategy by introducing spatiotemporal joint masking.

{\color{black}
In contrast, autoregressive methods predict future neural signals solely based on past observations, aligning with the brain's temporal integration mechanism, where current neural states largely depend on preceding activity. If a neural signal sequence is denoted as $X = (x_1, x_2, ..., x_T)$, an autoregressive BFM $f_{\theta}$ aims to predict the signal $x_t$ given the past $k$ observations $(x_{t-k}, ..., x_{t-1})$. The training objective typically involves minimizing the prediction error across the sequence, often defined as:
\begin{equation} \label{eq:autoreg_loss}
L_{\text{autoreg}} = \sum_{t=k+1}^{T} \mathcal{D}(x_t, f_{\theta}(x_{t-k}, ..., x_{t-1}))
\end{equation}
where $\mathcal{D}$ represents a suitable divergence or distance function (e.g., MSE for continuous signals or cross-entropy if signals are discretized into tokens). BrainLM \cite{carobrainlm}, for instance, employs a Transformer-based autoregressive framework for fMRI data, where the model sequentially predicts future time steps based on past neural activity. NeuroLM further advances this approach by introducing a multi-channel autoregressive mechanism, which integrates information across EEG channels to enhance joint modeling. Moreover, it incorporates a causal autoregressive framework, enforcing directional constraints in the prediction process, thereby improving the model's capability to capture EEG temporal patterns with greater fidelity.

\textbf{Remark:} \textit{Different modalities of brain data are suitable for different SSL strategies. Masked prediction is more suitable for modalities such as fMRI. These modalities have high spatial resolution, and brain regions often exhibit significant spatial co-activation patterns under the same cognitive state. By masking the signals of certain brain regions and requiring the model to perform reconstruction, the model can be guided to fully utilize the spatial dependencies in the data. Autoregression is more suitable for modalities with strong temporal continuity, such as EEG. These signals have high temporal resolution, reflecting the rapid dynamic changes in brain activity and exhibiting distinct temporal evolutionary characteristics. By predicting future signals based on historical time segments, the model can effectively capture the causal structure and temporal dependencies of neural activity.}

Contrastive Learning: It aims to construct positive and negative sample pairs to learn more discriminative representations by pulling similar samples closer while pushing dissimilar ones apart. This approach has been highly successful in visual and multimodal domains, notably exemplified by models like CLIP \cite{clip}, which aligns image and text representations. Many contrastive learning methods aim to maximize the agreement between different 'views' of the same underlying brain state or activity. Given an anchor sample $x_i$ (e.g., a segment of a brain signal), a positive sample $x_i^+$ (e.g., an augmented version of $x_i$ or a concurrently recorded related signal from another modality), and a set of $N-1$ negative samples $x_k$ (dissimilar signals), a common objective is the InfoNCE loss (or a variant):
\begin{equation} \label{eq:contrastive_loss}
L_{\text{contrastive}} = - \mathbb{E} \left[ \log \frac{\exp(\text{sim}(f_{\theta}(x_i), f_{\theta}(x_i^+)) / \tau)}{\exp(\text{sim}(f_{\theta}(x_i), f_{\theta}(x_i^+)) / \tau) + \sum_{k=1}^{N-1} \exp(\text{sim}(f_{\theta}(x_i), f_{\theta}(x_k)) / \tau)} \right]
\end{equation}
where $f_{\theta}$ is the BFM encoder that maps inputs to a representation space, $\text{sim}(\cdot, \cdot)$ denotes a similarity function (e.g., cosine similarity) between representations, and $\tau$ is a temperature hyperparameter. This loss encourages the model to learn representations that are invariant to certain augmentations (in intra-modality learning) or that align across different modalities (in brain-heterogeneous or inter-brain modality learning). Contrastive learning methods can generally be categorized into two types: \textbf{intra-brain modality contrastive learning} and \textbf{brain-heterogeneous modality contrastive learning}. The former focuses on contrastive objectives within brain data, either within the same modality or across different brain modalities. This approach facilitates the integration of multi-source neural signals and helps build more comprehensive representations of neural activity. The latter, on the other hand, involves contrasting brain data with data from heterogeneous modalities such as text or images. This strategy enables the representation of neural information in alternative modalities, thereby enhancing interpretability and broadening potential applications.
}


The core idea of intra-brain modality contrastive learning is to establish contrastive representations within the same modality while also capturing shared representations across different brain modalities, enabling the model to learn stable and comprehensive neural activity patterns. On one hand, within the same modality, despite variations in an individual’s physiological and psychological state, neural activity patterns exhibit temporal continuity and spatial consistency under the same task. For instance, while EEG fluctuations are influenced by neural oscillations, adjacent time segments still follow specific temporal dependencies, reflecting the stable functional organization of local neural networks. MBrain \cite{cai2023mbrain} constructs contrastive samples from different time segments across channels to enhance temporal stability, while EEGPT applies perturbation-based contrastive learning, introducing temporal jittering, frequency shifts, and channel permutation to generate semantically consistent but structurally varied EEG samples.
On the other hand, despite differences in acquisition methods, spatiotemporal resolution, and noise characteristics, shared neurodynamic relationships exist across different brain modalities due to cross-modal functional organization. For example, EEG captures rapid cortical electrical activity, whereas fMRI reflects slower neurovascular processes, yet high-frequency EEG oscillations (e.g., gamma waves) often correlate with fMRI BOLD signals during cognitive tasks. By aligning representations across EEG and fMRI, the model can extract shared information and establish consistent neural mappings. Brant-X \cite{zhang2024brant} introduces a two-level alignment mechanism, contrasting EEG with other physiological signals (ECG, EOG, EMG). Patch-Level alignment ensures that local physiological signals map to a shared representation space, while Sequence-Level alignment enforces consistency in long-term neural dynamics, enhancing cross-modal representation learning.

Brain-heterogeneous modality contrastive learning constructs cross-modal data pairs, allowing brain signals to transcend predefined labels and be expressed in richer modalities such as text or images. In traditional supervised learning, models categorize brain data within a fixed task framework, requiring explicit task labels. In contrast, contrastive learning removes this constraint, enabling models not only to encode neural signals but also to capture richer task-related contextual information. Through cross-modal mapping, the model learns task representations directly from brain signals, rather than being restricted to isolated task classifications.
This means that instead of simply distinguishing between labels in “motor imagery” or “emotion recognition” tasks, the model can associate brain activity with stimulus content (e.g., visual scenes or textual descriptions) to infer the task itself, leading to semantically richer neural representations. For example, Kong et al. \cite{kong2024toward} aligned fMRI signals with visual images using a contrastive learning framework. By employing the CLIP model, they optimized fMRI representations to match corresponding visual features, allowing the model not only to identify neural activation patterns but also to infer the visual stimuli perceived by the subject, demonstrating generalization to unseen individuals. Similarly, NeuroLM adopts a cross-modal contrastive strategy, aligning EEG signals with text representations. In the first stage, a discriminator evaluates EEG-text alignment, optimizing cross-modal correspondence through contrastive loss. In the second stage, multi-task instruction tuning further refines this alignment, allowing NeuroLM to generalize across diverse BCI tasks and enhance its adaptability to varied neural decoding scenarios.

{\color{black}
\textbf{Remark:} \textit{Although brain-heterogeneous modality contrastive learning currently offers limited improvements for direct brain data analysis, its long-term value cannot be overlooked. As model capabilities expand, the next phase of BFMs will place greater emphasis on flexible input-output modes to accommodate diverse brain data formats and task requirements. This strategy enables BFM to gradually align with current mainstream multimodal large language models, thereby acquiring the ability to handle complex semantic alignment and cross-modal understanding, providing the necessary foundational support for more advanced BCI tasks.}

Fine-tuning \& Task-Specific Adaptation: Once the BFM has been pre-trained using SSL and masking techniques, the model proceeds to the fine-tuning phase, where it is adapted for specific tasks using labeled data. The fine-tuning process involves supervised learning, where the model is trained to predict specific outcomes (e.g., brain disease diagnosis or cognitive state classification) based on the representations learned during pre-training.

Fine-tuning typically uses cross-entropy loss for classification tasks or mean squared error for regression tasks. Let $D=\left\{\left(x_i, y_i\right)\right\}$ represent the labeled dataset, where $x_i$ is the input brain signal, and $y_i$ is the corresponding label. The fine-tuning objective is to minimize the task-specific loss:

\begin{equation}
\mathcal{L}_{\text {fine-tune }}=\sum_i \mathcal{L}_{\text {task }}\left(f\left(x_i\right), y_i\right)
\end{equation}

For classification tasks, this might involve predicting discrete classes of brain activity or disease, while for regression tasks, the output might represent continuous values, such as cognitive scores or clinical measures.

\textbf{Remark:} \textit{Current exploration of fine-tuning in BFMs remains insufficient. Compared to text and images, brain data exhibits characteristics such as significant individual variability and unstable cross-task distribution, making BFMs more reliant on fine-tuning to achieve effective transfer. However, strategies such as unsupervised domain adaptation, test-time training, and efficient parameter fine-tuning methods (e.g., Adapter, LoRA) have not yet been widely applied in BFMs. By drawing on these mechanisms that have been validated as effective in NLP and CV, it is likely that the adaptability and practicality of BFMs in real-world BCI scenarios can be significantly enhanced.}

\textbf{Summary of Training Strategy:}
The suitability of SSL for brain data is uncertain due to high noise, variability, unlike NLP and CV data. These challenges hinder learning generalizable features with standard SSL, leading most current BFMs to rely heavily on extensive fine-tuning. This reliance curtails true cross-task generalization and zero-shot capabilities, even for models like NeuroLM. Furthermore, many BFMs training strategies are merely adaptations from NLP and CV, lacking neuroscience-specific optimizations. Consequently, masked prediction or contrastive learning often fails to incorporate vital biological constraints, functional connectivity, or complex brain dynamics. This positions current BFMs more as direct AI adaptations than neuro-tailored models, limiting their biological interpretability and real-world applicability.}

\section{Typical Applications of BFMs}

The potential of BFMs spans a wide range of applications, as shown in Fig. \ref{fig:applications}, which can be broadly categorized into Brain Decoding and Brain Discovery. These applications leverage the strengths of BFMs in both decoding brain activity and exploring the underlying mechanisms of brain function. Each category utilizes different configurations of BFMs, ranging from pretraining alone to the combination of pretraining with interpretability analysis, to address specific challenges in neural signal processing. These approaches provide a powerful framework for understanding and manipulating neural data in both clinical and research settings.

\subsection{Brain Decoding}










Brain decoding refers to using BFMs to interpret neural signals and decode mental states, intentions, or cognitive processes from brain activity. This process is critical for various applications, from BCIs to brain disease diagnostics. BFMs are particularly valuable in this domain due to their ability to generalize across diverse tasks, modalities, and scenarios, allowing for more robust and adaptable models for interpreting neural signals. Their flexibility makes them essential for advancing brain research and improving clinical and real-world applications. The typical BFMs for brain decoding are summarized in Tab. \ref{tab: summary} and Table \ref{tab:merged_bfm_comparison_centered}, which provide a quantitative performance comparison of representative models of fMRI and EEG data on public datasets, respectively.

\begin{table}[t]
\centering
\caption{\textcolor{black}{\scriptsize Summary of Typical BFMs for Brain Decoding.}
}
\vspace{-10pt}
\renewcommand{\arraystretch}{1} 
\setlength{\tabcolsep}{4pt}  
\resizebox{0.70\linewidth}{!}{
\begin{tabularx}{\textwidth}{
    >{\hsize=1.4\hsize\centering\arraybackslash}X     
    >{\hsize=0.75\hsize\centering\arraybackslash}X    
    >{\hsize=0.8\hsize\centering\arraybackslash}X    
    >{\hsize=1.15\hsize\centering\arraybackslash}X   
    >{\hsize=1.05\hsize\centering\arraybackslash}X   
    >{\hsize=1.15\hsize\centering\arraybackslash}X   
    >{\hsize=0.70\hsize\centering\arraybackslash}X    
@{}}
\toprule
\textbf{Model} & \textbf{Pretraining} & \textbf{Fine-tuning} & \textbf{Cross Scenarios} & \textbf{Cross Tasks} & \textbf{Multi Modalities} & \textbf{Size}\\
\midrule
BrainLM   \cite{carobrainlm}     & $\checkmark$ & $\checkmark$    & $\checkmark$ & $\checkmark$ & $\times$  &650M  \\
LaBraM   \cite{jiang2024large}      & $\checkmark$ & $\checkmark$ & $\checkmark$ & $\checkmark$ & $\times$  & 369M  \\
NeuroLM  \cite{jiang2024neurolm}      & $\checkmark$ & $\times$    & $\checkmark$ & $\checkmark$ & $\checkmark$ &1.7B \\
Brant   \cite{zhang2023brant}       & $\checkmark$ & $\checkmark$ & $\checkmark$ & $\checkmark$ & $\times$  &505M  \\
BrainSegFounder \cite{cox2024brainsegfounder} & $\checkmark$ & $\checkmark$ & $\times$     & $\times$     & $\checkmark$ &69M\\
MeTSK   \cite{cui2023meta}       & $\checkmark$ & $\checkmark$ & $\times$     & $\checkmark$ & $\times$  &---  \\
AnatCL   \cite{barbano2024anatomical}      & $\checkmark$ & $\times$    & $\checkmark$ & $\checkmark$ & $\times$  &---  \\
BRAINBERT  \cite{sihag2024towards}    & $\checkmark$ & $\checkmark$ & $\checkmark$ & $\checkmark$ & $\times$  &43M  \\
NeuroVNN  \cite{sihag2024towards}    & $\checkmark$ & $\checkmark$ & $\checkmark$ & $\times$   & $\times$  &---  \\
Brain-JEPA  \cite{dong2025brain}   & $\checkmark$ & $\checkmark$ & $\times$ & $\checkmark$ & $\times$ & 307M  \\
FoME  \cite{shi2024fome}    & $\checkmark$ & $\checkmark$ & $\checkmark$ & $\checkmark$ & $\times$  &745M  \\
FM-BIM  \cite{karimi2024approach}   & $\checkmark$ & $\checkmark$ & $\times$     & $\checkmark$ & $\times$  &---  \\
BrainWave  \cite{yuan2024brainwave}    & $\checkmark$ & $\checkmark$ & $\checkmark$ & $\checkmark$ & $\checkmark$  & ---  \\
TF-C  \cite{zhang2022self}   & $\checkmark$ & $\checkmark$ & $\times$ & $\checkmark$ & $\checkmark$ &---   \\
Neuro-GPT  \cite{cui2024neuro}  & $\checkmark$ & $\checkmark$ & $\checkmark$ & $\times$ & $\times$  &80M  \\
BrainMAE   \cite{yang2024brainmae}   & $\checkmark$ & $\checkmark$ & $\checkmark$ & $\checkmark$ & $\times$  &---  \\
EEGFormer  \cite{chen2024eegformer}   & $\checkmark$ & $\checkmark$ & $\times$ & $\checkmark$ & $\times$  &---  \\
CEReBrO  \cite{dimofte2025cerebro}   & $\checkmark$ & $\checkmark$ & $\times$ & $\checkmark$ & $\times$  &85M  \\
CBraMod  \cite{wang2024cbramod}   & $\checkmark$ & $\checkmark$ & $\checkmark$ & $\checkmark$ & $\times$  &4M  \\
EEGPT-Yue  \cite{yue2024eegpt}    & $\checkmark$ & $\checkmark$ & $\checkmark$ & $\checkmark$ & $\times$   &1.1B \\
TGBD  \cite{kong2024toward}   & $\checkmark$ & $\checkmark$ & $\checkmark$ & $\times$ &$\times$  &--- \\
MBrain  \cite{cai2023mbrain}    & $\checkmark$ & $\checkmark$ & $\checkmark$ & $\times$ & $\checkmark$   &--- \\
EEGPT   \cite{wang2025eegpt}   & $\checkmark$ & $\checkmark$ & $\times$ & $\checkmark$ & $\times$   &10M \\
Brant-X  \cite{zhang2024brant}   & $\checkmark$ & $\checkmark$ & $\times$ & $\checkmark$ & $\checkmark$  &1B \\
FM-APP  \cite{he2024fm}  & $\checkmark$ & $\checkmark$ & $\times$ & $\checkmark$ & $\checkmark$  &---\\

\bottomrule
\end{tabularx}}
\label{tab: summary}
\begin{tablenotes}
\item \textcolor{black}{\scriptsize "Cross scenarios": the ability of BFMs to operate effectively across diverse settings and conditions within a single paradigm; \item "Cross tasks capability": BFMs can achieve robust performance across different paradigms.}
\end{tablenotes}
\end{table}

\subsubsection{Pretraining Only}
The Pretraining Only approach trains BFMs on large, diverse datasets of neural signals without specific task adaptation. The model learns universal representations that capture core patterns of brain activity shared across various contexts, scenarios, and experimental conditions. This makes it highly applicable in cross-scenario tasks, where neural data may vary across a wide range of conditions, such as: \textbf{1) Epilepsy Detection Across Multiple Scenarios:} In epilepsy diagnosis, BFMs pretrained on EEG data from diverse seizure scenarios, such as different seizure types, environmental conditions, and patient populations, can generalize across different seizure manifestations. By leveraging these generalized features, the pretrained model can effectively recognize seizures in new datasets that may differ in subjects, recording settings, or seizure types. This capability is crucial for developing real-time seizure detection systems for clinical use. \textbf{2) Wide Range Speech-BCI Data Decoding:} For speech-evoked BCI systems, BFMs pretrained on diverse speech-related brain data captured from different subjects, speech tasks, and sensory modalities can decode speech intentions more effectively across multiple scenarios. Whether the data comes from speech imagery, silent speech, or actual speech, the pretrained model can handle this variability and decode speech-related mental states. This application is critical for improving the accessibility of communication devices for individuals with speech impairments.



\subsubsection{Pretraining + Fine-tuning}
The Pretraining + Fine-tuning approach combines the power of pretrained models with task-specific adaptation. After pretraining on large-scale datasets, BFMs are fine-tuned for specific tasks, enhancing their performance in specialized applications. This approach is particularly effective in cross-task scenarios, where the same model needs to be adapted for multiple different types of BCI tasks. This flexibility is advantageous for applications that require the model to work across a range of BCI paradigms, such as: \textbf{1) Cross-Task BCI Applications:} A model pretrained on diverse neural data can be fine-tuned for various BCI tasks. For example, a single model can be adapted for motor imagery, cognitive workload assessment, and emotion recognition. This enables a unified framework capable of handling multiple BCI applications without task-specific model design, making the model more versatile and reducing development time and costs. \textbf{2) Generalist Disease Diagnosis:} Pretraining on general neural signal data, followed by fine-tuning for specific diseases (e.g., PTSD, anxiety, and neuroticism), enables BFMs to identify biomarkers for disease diagnosis and progression. This approach can be particularly useful for developing diagnostic tools that can work across a wide range of patients, reducing the need for collecting large amounts of task-specific training data. By leveraging the generalized knowledge from pretraining, the fine-tuned model can provide accurate and reliable predictions for various diseases based on neural data.


\begin{figure}
    \centering
    \includegraphics[width=0.70\linewidth]{image/appfinal2.pdf}
    \vspace{-10pt}
    \caption{\scriptsize This diagram demonstrates two primary application domains of BFMs. Two paradigms in the Brain Decoding category are highlighted: (A.a) Cross Scenarios, which involve different subtasks or variations within the same task. For example, detecting local and global epilepsy with one BFM falls under Cross Scenarios, as does performing speech-BCI with different languages. (A.b) Cross Tasks encompass entirely distinct tasks, such as emotion recognition, sleep monitoring, and fatigue assessment. Similarly, using one BFM to diagnose conditions like PTSD, anxiety, and neuroticism also falls under the Cross Tasks category. In contrast, the (B) Brain Discovery domain focuses on constructing digital twin brains using BFMs. Through interpretability analysis of these digital models, researchers can simulate and explore the underlying mechanisms of the biological brain.}
    \label{fig:applications}
\end{figure}

\begin{table}[ht]
\centering
\caption{\scriptsize\textcolor{black}{Comparison of Typical fMRI and EEG based BFM models.  MSE (Mean Squared Error), $\rho$ (Pearson Correlation Coefficient), ACC (Accuracy), F1 (F1 Score), AUROC (Area Under the Receiver Operating Characteristic Curve).}}
\label{tab:merged_bfm_comparison_centered}
\resizebox{0.75\textwidth}{!}{%
\begin{tabular}{@{}ccccccc@{}} 
\toprule
Dataset      & Task                       & Metrics & BrainLM\cite{carobrainlm}           & Brain-JEPA\cite{dong2025brain}        & NeuroLM \cite{jiang2024neurolm}           & LaBraM\cite{jianglarge}            \\ \midrule
\multicolumn{7}{c}{\textit{fMRI-based Models}} \\ \midrule 
\multirow{8}{*}{HCP-Aging} & Age & MSE & $0.331 \pm 0.018$ & $0.298 \pm 0.017$ & ---               & ---               \\
             & Age                        & $\rho$  & $0.832 \pm 0.028$ & $0.844 \pm 0.030$ & ---               & ---               \\
             & Sex                        & ACC     & $74.39 \pm 1.55$  & $81.52 \pm 1.03$  & ---               & ---               \\
             & Sex                        & F1      & $77.51 \pm 1.13$  & $84.26 \pm 0.82$  & ---               & ---               \\
             & Neuroticism                & MSE     & $0.942 \pm 0.082$ & $0.897 \pm 0.055$ & ---               & ---               \\
             & Neuroticism                & $\rho$  & $0.231 \pm 0.012$ & $0.307 \pm 0.006$ & ---               & ---               \\
             & Flanker                    & MSE     & $0.971 \pm 0.054$ & $0.972 \pm 0.038$ & ---               & ---               \\
             & Flanker                    & $\rho$  & $0.318 \pm 0.048$ & $0.406 \pm 0.027$ & ---               & ---               \\ \midrule
\multirow{4}{*}{ADNI}      & NC/MCI                     & ACC     & $75.79 \pm 1.05$  & $76.84 \pm 1.05$  & ---               & ---               \\
             & NC/MCI                     & F1      & $85.66 \pm 1.27$  & $86.32 \pm 0.54$  & ---               & ---               \\
             & Amyloid $\alpha\beta\,\text{+ve/-ve}$ & ACC & $67.00 \pm 7.48$  & $71.00 \pm 4.90$  & ---               & ---               \\
             & Amyloid $\alpha\beta\,\text{+ve/-ve}$ & F1  & $68.82 \pm 8.48$  & $75.97 \pm 3.93$  & ---               & ---               \\ \midrule
\multicolumn{7}{c}{\textit{EEG-based Models}} \\ \midrule 
\multirow{2}{*}{TUAB}      & Abnormal Detection         & ACC     & ---               & ---               & $0.7969 \pm 0.0091$ & $0.8140 \pm 0.0019$ \\
             & Abnormal Detection         & AUROC   & ---               & ---               & $0.7884 \pm 0.0194$ & $0.9022 \pm 0.0009$ \\ \midrule
\multirow{2}{*}{TUEV}      & Event Type Classification  & ACC     & ---               & ---               & $0.4679 \pm 0.0356$ & $0.6409 \pm 0.0065$ \\
             & Event Type Classification  & F1 Score& ---               & ---               & $0.7359 \pm 0.0219$ & $0.8312 \pm 0.0052$ \\ \midrule
\multirow{2}{*}{SEED}      & Emotion Recognition        & ACC     & ---               & ---               & $0.6034 \pm 0.0010$ & $0.7318 \pm 0.0019$ \\
             & Emotion Recognition        & F1 Score& ---               & ---               & $0.6063 \pm 0.0030$ & $0.7354 \pm 0.0021$ \\ \bottomrule
\end{tabular}
}
\end{table}

\subsection{Brain Simulation and Discovery}

Brain discovery extends beyond decoding brain activity to simulating and understanding the brain’s underlying biological mechanisms. This field is crucial for uncovering fundamental aspects of brain function, by building digital brain models that mimic biological brain activity. These models allow researchers to explore and interpret how various neural networks interact during cognitive processes and disease progression. In this context, the primary goal is to construct a digital brain, a simulation of the biological brain that can be studied and analyzed to discover previously unknown aspects of neural function.

{\color{black}
\subsubsection{Pretraining + Interpretability}
In the Pretraining + Interpretability approach, BFMs are pretrained on large-scale neural signal datasets (e.g., EEG, fMRI) to learn general patterns of brain activity. These models are then fine-tuned and structurally constrained using multimodal brain atlases (e.g., Brainnetome) to construct a Digital Twin Brain (DTB)~\cite{xiong2023digital}. The modeling pipeline integrates a multiscale architecture: the microscopic level models individual neurons with detailed electrical dynamics (e.g., Hodgkin–Huxley, integrate-and-fire); the mesoscopic level captures interactions within and between excitatory and inhibitory populations (e.g., Wilson-Cowan, dynamic mean-field); and the macroscopic level simulates large-scale activity and inter-regional functional connectivity using whole-brain network models. Training aims to reproduce empirical functional connectivity and signal variability via model inversion or parameter exploration. Validation is performed by comparing simulated outputs with neural data and assessing responses to perturbation-based simulations.


Through interpretability techniques such as perturbation analysis or attention mechanisms, researchers can analyze the digital brain and gain insights into how the biological brain operates under various conditions. This approach is particularly valuable for exploring complex neural dynamics and uncovering mechanisms that are difficult to observe directly. Key applications include: \textbf{1) Exploring Brain Function and Disease Mechanisms:} The digital brain serves as a platform for studying brain function in both healthy and diseased states. Perturbation analysis can simulate disruptions such as those from neurodegenerative diseases (e.g., Alzheimer's, Parkinson’s), allowing researchers to examine network-level effects, understand disease progression, and identify potential biomarkers or therapeutic targets. \textbf{2) Understanding Cognitive Processes:} Digital brain models enable simulation of complex cognitive functions like memory, decision-making, and learning. By examining interactions among brain regions, researchers can uncover neural mechanisms underlying cognition and identify regions linked to cognitive disorders, aiding diagnosis and targeted treatment. \textbf{3) Uncovering Unknown Mechanisms in the Biological Brain:} DTB models offer access to neural processes difficult to observe in vivo. Interpretability methods help reveal which regions or networks are critical for specific cognitive or emotional functions, offering new insights into behavior, mental health, and brain plasticity.
}

{\color{black}
\section{Future Directions and Challenges}
As BFMs continue to evolve, several challenges and opportunities remain in their data, model, training, application, and interpretability. This section discusses several future directions and the corresponding challenges in advancing BFMs for neuroscience, neuroimaging, and clinical applications.

\subsection{Data for Brain Foundation Models}
Data availability, diversity, and quality are foundational to the success of BFMs. Future research should prioritize the integration of diverse brain imaging modalities such as fMRI for dynamic activity patterns, EEG and magnetoencephalography (MEG) for high temporal resolution, diffusion tensor imaging (DTI) for structural connectivity, and structural MRI for anatomical mapping. Incorporating behavioral assessments, clinical labels, and genetic information will further enable more holistic modeling of brain function and disorders. To address limited data coverage across populations and conditions, generative approaches such as GANs or diffusion models can be explored to synthesize realistic brain signals, particularly for rare diseases and underrepresented cohorts. 

However, key challenges remain. The variability in imaging protocols, scanner configurations, and population demographics leads to inconsistent datasets that reduce model generalizability. The absence of standardized preprocessing and annotation pipelines further complicates data harmonization and reproducibility. Addressing these issues requires the development of community-adopted protocols for cross-site harmonization, along with publicly available benchmarks for evaluating BFM performance. Moreover, technical strategies to meet data protection regulations such as GDPR and HIPAA must be tightly integrated into model design and deployment.

\subsection{Architectural Innovations and Neuroscience Integration}
BFMs face unique architectural challenges due to the brain’s highly structured, multimodal, and dynamic nature. While models such as CNNs, Transformers, and RNNs offer a foundation, they must be adapted to reflect domain-specific constraints. A key research direction involves integrating neuroscientific priors directly into model design. For instance, brain connectivity matrices derived from DTI or functional connectivity analyses can be used to define the topologies of Graph Neural Networks (GNNs), encoding structural or functional constraints as part of the model architecture. This approach enables spatial relationships between brain regions to be explicitly modeled, rather than inferred indirectly.

Beyond structural priors, models should also be encouraged to reflect known functional hierarchies and spectral dynamics. One promising direction is the design of loss functions that incorporate neuroscientific principles, such as promoting representational consistency with known brain oscillations (e.g., alpha, beta, and gamma bands) or penalizing divergence from established cortical gradients. Incorporating these constraints can guide model learning toward biologically plausible representations and may improve transferability across tasks and populations.

\subsection{Training Strategies and Multimodal Fusion}
Training BFMs effectively requires learning paradigms that can extract meaningful patterns from vast quantities of unlabeled data while respecting the constraints of biological realism. SSL and contrastive learning methods are particularly suitable, as they reduce dependence on manual annotations while capturing rich neural dynamics. These methods must be adapted to account for domain-specific objectives, such as predicting inter-regional correlations or simulating signal transitions across cognitive states. Transfer learning and meta-learning strategies can further help BFMs generalize across different populations and tasks with limited fine-tuning.

A major research direction involves the development of unified training frameworks that integrate data from multiple modalities, such as EEG, fNIRS, fMRI, MEG, and even genetics, into a coherent representational space. Aligning these modalities poses technical challenges due to their differences in spatial and temporal resolution, data scale, and noise characteristics. Advances in multimodal fusion techniques, such as cross-modal encoders or shared latent spaces, are needed to enable effective integration. Furthermore, training costs must be addressed through hardware-efficient strategies, such as distributed computing and lightweight architectures, to ensure the scalability and accessibility of BFMs for both academic and clinical settings.

\subsection{Adaptation and Use of Pretrained BFMs}
Once pretrained, BFMs offer significant potential for downstream applications, including early disease diagnosis, cognitive modeling, and BCI development. Fine-tuning these models for specific clinical or research tasks requires protocols that are both data-efficient and modality-aware. Emerging approaches such as prompt tuning or adapter modules could provide flexible, parameter-efficient strategies for adaptation without extensive retraining.

In parallel, integrating BFMs with LLMs opens new opportunities for developing interpretable, interactive neuro-AI systems. These systems could interpret complex brain signals and provide natural language explanations for predictions, thereby supporting clinical decision-making or cognitive assessments. Nonetheless, several challenges need to be resolved, including ensuring robust domain adaptation across datasets and populations, maintaining low-latency inference for real-time applications like BCIs, and ensuring the trustworthiness of model outputs through validation mechanisms grounded in neuroscientific principles.

\subsection{Interpretability and Evaluation Standards}
Interpretability is a critical requirement for the acceptance and safe deployment of BFMs in neuroscience and clinical contexts. Traditional deep learning models often act as black boxes, making it difficult to understand or validate their decisions. Future work must focus on designing interpretability techniques that are both technically rigorous and neuroscientifically meaningful. This includes developing attention mechanisms and saliency-based visualizations that highlight relevant brain regions, as well as concept-based approaches that map predictions to known cognitive or clinical constructs.

To move beyond qualitative insights, the field needs standardized interpretability metrics specifically tailored to brain modeling. These could include overlap with task-evoked activation maps, correspondence with known anatomical regions, or alignment with expert annotations. Neuro-symbolic models that combine statistical learning with interpretable rules derived from brain science offer another promising direction. Moreover, combining BFMs with multimodal language models could help translate activation patterns into human-readable explanations, fostering stronger collaboration between AI systems and clinicians or neuroscientists.

\section*{Conclusion}

In this survey, we introduce the concept of BFMs as a transformative paradigm in neural signal processing and brain discovery. By providing the first formal definition and theoretical framework for BFMs, we delineate how they differ from conventional foundation models, emphasizing their specialized design for handling complex, noisy, and heterogeneous neural data.
We synthesize the latest architectural innovations and offer novel perspectives on their applications. Our analysis highlights that BFMs not only extend the frontiers of brain decoding by generalizing robustly across diverse scenarios and tasks but also enable advanced brain simulation and discovery through digital twin brain construction. This dual capability underscores their potential to revolutionize both clinical diagnostics and neuroscientific research.
Furthermore, we outline a forward-looking roadmap identifying critical challenges and future directions. These include improving data integration and quality, developing novel training strategies and multimodal architectures, and enhancing model interpretability. Overcoming these challenges is essential to fully harness the transformative potential of BFMs in personalized medicine, neurorehabilitation, and beyond.
Ultimately, this survey serves as a foundational reference for researchers, practitioners, and policymakers. By bridging the gap between neuroscience and artificial intelligence, BFMs are poised to reshape our understanding of brain function and drive the development of next-generation brain technologies.
}

\bibliographystyle{IEEEtran}
\bibliography{ref}





%
%
%
%
%
\section*{Biographies}
\label{sec:bio}
\vspace{-70pt}
%
%

{\tiny

\begin{IEEEbiographynophoto}
{Xinliang Zhou}
received the B.E. degree in Mechatronics Engineering from Beijing Jiaotong University, Beijing, China, in 2021, and the Ph.D. degree in Computer Science from Nanyang Technological University, Singapore, in 2025. His research interests include Brain-Computer Interfaces, Brain Foundation Models, and Interpretable Artificial Intelligence. (Email Address: xinliang001@e.ntu.edu.sg)
\end{IEEEbiographynophoto}
\vspace{-65pt} 

\begin{IEEEbiographynophoto}{Chenyu Liu} received the B.E. degree in Software Engineering from the University of Electronic Science and Technology of China, Sichuan, China, in 2020. He is currently working toward the Ph.D. degree with the College of Computing and Data Science, Nanyang Technological University, Singapore. His research interests include Brain-Computer Interface, Multivariate Time Series, and Multimodal Spatio-Temporal Data Mining. (Email Address: chenyu003@e.ntu.edu.sg)
\end{IEEEbiographynophoto}

\vspace{-65pt}

\begin{IEEEbiographynophoto}{Zhisheng Chen} received the B.E. degree in Remote Sensing Science and Technology from China University of Mining and Technology. He is currently working toward the Master's degree with the  Institute of Computing Technology, Chinese Academy of Sciences. His research interests include Brain-Computer Interfaces, Large-scale Neural Signal Modeling, and  Brain Foundation Models. (Email Address: zhishengchen@cumt.edu.cn)
\end{IEEEbiographynophoto}

\vspace{-65pt}

\begin{IEEEbiographynophoto}{Kun Wang} received the Ph.D. degree in Computer Science from University of Science and Technology of China (USTC), Anhui, China, in 2024. He is a Research Fellow at the College of Computing and Data Science, Nanyang Technological University. His primary research interests include ML4Science, Foundation Models and AI Agents. (Email Address: wang.kun@ntu.edu.sg)
\end{IEEEbiographynophoto}


\vspace{-60pt}

\begin{IEEEbiographynophoto}{Ziyu Jia} received the Ph.D. degree in Computer Science and Technology from Beijing Jiaotong University, Beijing, China, in 2022. He is an Assistant professor at the Brainnetome Center, Institute of Automation, Chinese Academy of Sciences. His interests include Novel Time Series Data Mining, Machine Learning, and Deep Learning Algorithms. He is also interested in various physiological time series applications in affective Brain-Computer Interfaces and Sleep Stage Classification. (Email Address: jia.ziyu@outlook.com)
\end{IEEEbiographynophoto}

\vspace{-60pt}

\begin{IEEEbiographynophoto}{Yi Ding} earned a Ph.D. in Computer Science and Engineering from Nanyang Technological University, Singapore, in 2023, a master's degree in Electrical and Electronics Engineering from the same university in 2018, and a bachelor's degree in Information Science and Technology from Donghua University, Shanghai, China, in 2017. He is a Research Assistant Professor at the College of Computing and Data Science, Nanyang Technological University. His research interests encompass Brain-Computer Interfaces, Deep Learning, Neural Signal Processing, Foundation Models, and Affective Computing. (Email Address: ding.yi@ntu.edu.sg)
\end{IEEEbiographynophoto}

\vspace{-60pt}


\begin{IEEEbiographynophoto}{Qingsong Wen} is Head of AI \& Chief Scientist at Squirrel AI Learning. He previously held positions at Alibaba, Qualcomm, and Marvell, and earned his M.S. and Ph.D. in Electrical and Computer Engineering from the Georgia Institute of Technology. His research focuses on Machine Learning, Data Mining, and Signal Processing, particularly in AI for Time Series (AI4TS), AI for Education (AI4EDU), LLMs, Generative AI, and AI Agents. He serves as Area Chair for conferences including NeurIPS and KDD, Associate Editor for IEEE TPAMI, IEEE SPL, and Neurocomputing. He is a member of the IEEE Machine Learning for Signal Processing Technical Committee (MLSP TC). (Email Address: qingsongedu@gmail.com)
\end{IEEEbiographynophoto}
}

\vfill

\end{document}